\documentclass[11pt]{article}
\usepackage[margin=1.1in]{geometry}
\usepackage{amsmath,amssymb,amsthm}
\usepackage{graphicx}
\usepackage{booktabs}
\usepackage{float}
\usepackage{hyperref}
\usepackage{microtype}
\usepackage{xspace}
\usepackage{xcolor}
\usepackage{caption}

\hypersetup{colorlinks=true,linkcolor=blue!50!black,citecolor=blue!50!black,urlcolor=blue!50!black}

\graphicspath{{outputs/figures/final/}}

\title{Structural Grid Descriptors Predict Within-Task Solver Success on ARC-AGI}
\author{Ayan Pendharkar}
\date{2026}

\newcommand{\tauhalf}{\ensuremath{\tau=0.5}\xspace}

\begin{document}
\maketitle

\begin{abstract}
We ask whether structural properties of intermediate grid states predict whether a symbolic
ARC-AGI solver will succeed. We frame this as a test of conditional mutual information
$I(X;Y\mid \mathrm{task}) > 0$. Across 44{,}800 runs spanning two architecturally distinct solvers
(beam search and Stochastic DFS), 400 ARC tasks, 28 configurations per solver, and both the ARC
training and evaluation splits, hand-crafted grid descriptors measured at the 50\% trajectory
fraction discriminate successful from failed runs within the same task (mean within-task best-feature
AUC $=0.885$, $p<0.001$ under within-task label permutation). The large run count is for coverage. The
within-task analyses are driven by a subset of tasks with both successes and failures, on the order of
30 to 40 reliable tasks per solver, and we report effective sample sizes throughout. Most of the
predictive content lies along a single grid-complexity axis (the descriptor space is near rank-1), so the
honest version of the claim is that one interpretable complexity descriptor predicts success rather than a
broad family of features. The signal generalizes across solvers:
a feature frozen on one solver architecture predicts success on the other at AUC $0.747$ to $0.762$
in all four transfer directions ($p<0.001$, leakage-controlled, with the feature selected on the source
solver only). On a pre-registered held-out set of 41 reliable diverse tasks, the frozen feature
\texttt{n\_components\_final} reaches AUC $=0.765$ (95\% CI $[0.717, 0.810]$, $p<0.001$), and the result
holds under a task-clustered bootstrap and when tasks shared across solvers are collapsed or dropped. The
held-out value is below the in-sample $0.885$, as expected once feature selection is removed. The signal is
not explained by solver
capacity (configuration-residualized AUC $=0.927$ on beam search and $0.896$ on SDFS, $p<0.001$), and
within tasks it is only weakly coupled to the score trajectory ($R^2 \approx 0$). As an application, an early-stopping
rule at the 50\% trajectory fraction saves 33.6\% of beam-search compute at 98.9\% solve-rate retained,
roughly ten times better than random. On SDFS, detecting degenerate trajectories saves 65.3\% of compute
at zero solve-rate cost. Finally, we examine the mechanism behind those degenerate SDFS trajectories:
on 229 of 400 evaluation tasks the DSL primitive library does not produce any valid transition from the
input grid. This 0-step collapse is budget-invariant and is also universally failed by beam search, which
is consistent with a DSL coverage limitation rather than a search-budget effect.
\end{abstract}

\section{Introduction}

The Abstraction and Reasoning Corpus (ARC-AGI) poses tasks that require inferring a transformation rule
from a handful of input--output grid examples and applying it to a held-out input. A common class of
solvers searches over a domain-specific language (DSL) of grid transformations, scoring candidate grids
against the target and expanding the most promising ones. These searches generate \emph{trajectories}:
sequences of intermediate grid states between the input and a final candidate.

This paper asks a narrow question about those trajectories. Holding the task fixed, does the
\emph{structure} of an intermediate grid carry information about whether the run that produced it will
succeed? By structure we mean simple, measurable properties: the number of objects a grid contains, the
entropy of its color distribution, the count of connected components. We refer to these throughout as
structural \emph{descriptors}. We avoid the word ``invariant'' deliberately: these are not quantities
conserved under the DSL or symmetry invariants in any mathematical sense, but hand-crafted structural
statistics evaluated at a fixed fraction of the trajectory. Writing $X$ for a
vector of such descriptors measured at a fixed point in the trajectory and $Y$ for the binary success
outcome, we test whether the conditional mutual information $I(X;Y\mid\mathrm{task})$ is greater than zero.
This is a within-task question by construction. It asks whether $X$ separates successes from failures
\emph{among runs on the same task}, not whether it tracks how hard different tasks are.

The distinction between a within-task signal and a cross-task one is the central conceptual point of the paper, so it is worth stating plainly. A feature can look strongly predictive across the whole dataset for an uninteresting reason: harder tasks tend to have both distinctive grids and lower success rates, so the feature ends up tracking task difficulty rather than telling us anything about which run on a given task will succeed. That is the cross-task signal, and it is mostly a restatement of task identity. The
within-task signal is the harder and more useful thing. It asks, once the task is fixed, whether the structure of an intermediate grid still separates the runs that succeed from the runs that fail. Our analyses are built to isolate the second from the first.

The within-task framing matters because the obvious confounds in ARC predictive analyses are all cross-task. Harder tasks have both distinctive grids and lower success rates, so a feature can look predictive across the whole dataset while carrying no information about which run on a given task will succeed. To control for this, we measure every effect within tasks, and we report significance under permutation tests that shuffle outcome labels \emph{within each task}, which leaves the cross-task structure intact.

The path to this result was not a straight line, and we think the route is worth stating because it is the reason to trust the destination. We did not set out to show that structural features predict success. We set out to ask whether they \emph{fail} to, and we spent most of the project trying to explain the within-task signal away. At first, it looked like it might be an artifact of score saturation, since the score trajectory flattens near the end of a run. When we controlled for that, the signal remained, so we asked whether it was simply task identity in disguise. When the within-task and task-normalization analyses ruled that out, we asked whether it was solver capacity, since more powerful configurations succeed more often. When residualization ruled that out, we asked whether it was specific to one search algorithm. Each time we tried to account for the effect, it remained, and the cross-solver transfer was the last of these checks. We report the convergent controls in that spirit: they are the alternative explanations we could not make stick, not a list of confirmations we sought out.
Section~\ref{sec:narrative} lays out this sequence explicitly so that the controls read as deliberate elimination rather than post-hoc selection.

With that framing, the evidence falls into four pieces:
\begin{enumerate}
  \item \textbf{Within-task signal.} Thirteen structural descriptors at \tauhalf{} discriminate success
  from failure within tasks at mean AUC $=0.885$ ($p<0.001$), and the effect appears for both solvers and
  both ARC splits.
  \item \textbf{Cross-solver generalization (our main result).} A feature frozen on one solver
  predicts success on the other at AUC $0.747$ to $0.762$, $p<0.001$, in all four transfer directions, leakage-controlled. Because the predictor is fixed on one architecture and tested on another, this result is the least attributable to a single search process, and we treat it as the central finding.
  \item \textbf{Held-out replication.} The pre-registered frozen feature \texttt{n\_components\_final}
  reaches AUC $=0.765$ (95\% CI $[0.717, 0.810]$, $p<0.001$) on 41 reliable diverse held-out tasks, and the result holds under a task-clustered bootstrap and when the tasks shared across solvers are collapsed or
  dropped (Section~\ref{sec:heldout}).
  \item \textbf{Confound control.} The signal survives capacity de-confounding (configuration-residualized AUC $=0.927$ beam, $0.896$ SDFS, $p<0.001$), and within tasks it is only weakly coupled to the score trajectory ($R^2\approx 0$).
\end{enumerate}

Those four pieces are the main result. The rest of the paper builds on it in two subordinate ways. As an
\emph{application}, an early-stopping rule that terminates runs predicted to fail at \tauhalf{} saves
33.6\% of beam-search compute while retaining 98.9\% of solves, roughly an order of magnitude better than randomly stopping the same fraction of runs. As a \emph{mechanistic observation}, we examine a failure mode we encountered along the way: on 229 of 400 evaluation tasks SDFS produces degenerate zero-step trajectories. The evidence is consistent with a DSL coverage gap, where the primitive library cannot produce any valid transition from the input grids on those tasks, and we show that the effect is budget-invariant and shared with beam search. Detecting these degenerate runs saves 65.3\% of SDFS compute at zero solve-rate cost. We present both as consequences of the main finding rather than as independent contributions.

\section{Related Work}

\paragraph{Predicting search success in program synthesis.}
The closest prior work is BUSTLE \cite{odena2020bustle}, which learns a neural model to predict whether intermediate values lie on the path to a correct program in bottom-up string-manipulation synthesis, and uses those predictions to guide search. Our question is the same in spirit, namely whether intermediate state predicts eventual success, but the setting and method differ in three ways. We study grid transformations rather than string DSLs. We use a small set of interpretable, hand-crafted structural descriptors rather than a learned neural classifier. And our analysis is post-hoc and confound-controlled, designed to establish \emph{whether} the signal exists and \emph{how much} of it survives explicit controls for task identity and solver capacity, rather than to maximize end-to-end solve rate.

\paragraph{Human and conceptual structure in ARC.}
H-ARC \cite{legris2025harc} collects human problem-solving trajectories on ARC, which gives a complementary view of search dynamics: where we characterize the trajectories of symbolic solvers, H-ARC characterizes those of people. ConceptARC \cite{moskvichev2023conceptarc} organizes ARC-style tasks by the abstract concept each one requires, and documents the concept-level abstraction that makes ARC hard for both machines and, in systematic ways, humans. That work motivates why structural grid features are only a partial signal: they describe the surface form of an intermediate grid, not the latent concept a task encodes.

\section{Methods}

\subsection{Two solvers}

Both solvers operate over the same \texttt{arc-dsl} transform library and the same scoring function
$S(g,t) = \mathrm{matches}/(H\times W)$, the pixel-match fraction between a grid $g$ and the target $t$. A
run succeeds if and only if $S=1.0$. The solvers differ in search architecture.

\paragraph{Solver A, beam search.}
At each step the solver applies all primitives to all current beams, scores each resulting grid against
the target, and keeps the top $K$. We use 28 configurations that vary beam width (2 to 32), score noise
$\sigma\in\{0, 0.05, 0.10\}$, and primitive set (full, geometric, or no-scaling), with multiple seeds on
the noisy configurations to produce within-task outcome diversity. The configuration capacity score is
$\log_2(\text{beam width}) + 20\sigma + p$, where $p=0$ for the full primitive set and $p=-0.3$ for the
geometric or no-scaling subsets.

\paragraph{Solver B, Stochastic DFS (SDFS).}
SDFS is randomized iterative-deepening depth-first search with stochastic transform ordering. It is
architecturally distinct from beam search: it commits to a single active path and backtracks on budget
exhaustion (depth-first), rather than maintaining $K$ concurrent paths explored level by level
(best-first). Configurations are either deterministic (alphabetical transform order with backtracking) or
stochastic (independent random-restart walks from the root, with
$\text{n\_restarts} = \text{max\_nodes} // \text{max\_depth}$). The 28 configurations comprise 8
deterministic runs plus 4 stochastic settings $\times$ 5 seeds, with \texttt{max\_nodes} ranging from 32
to 512 and capacity score $\log_2(\text{max\_nodes}) + p$ (same penalty structure as Solver A). The
important detail is that SDFS \texttt{best\_steps} are materialized into the same schema as beam search
(\texttt{step[grid]}, \texttt{step[score]}), so the feature extractor is unchanged across the two solvers.

Each solver is run on all 400 ARC tasks across both the training and evaluation splits, 28 configurations
each, for a total of 44{,}800 runs. Table~\ref{tab:data} summarizes the collection.

\begin{table}[H]
\centering
\caption{Data collection. ``Reliable diverse'' tasks satisfy the pre-registered floor
$n_{\text{pos}}\geq 3$ and $n_{\text{neg}}\geq 3$.}
\label{tab:data}
\small
\begin{tabular}{lrrrrr}
\toprule
Run & Tasks & Runs & Successes & Diverse & Reliable diverse \\
\midrule
Solver A training (Stage 1) & 400 & 11{,}200 & 867 (7.7\%) & 34 & 21 \\
Solver B training           & 400 & 11{,}200 & 301 (2.7\%) & 38 & 29 \\
Solver A evaluation         & 400 & 11{,}200 & 270 (2.4\%) & 22 & 14 \\
Solver B evaluation         & 400 & 11{,}200 &  87 (0.8\%) & 18 &  8 \\
\midrule
Solver A pooled             & 800 & 22{,}400 & ---         & 56 & 35 \\
Solver B pooled             & 800 & 22{,}400 & ---         & 56 & 37 \\
\bottomrule
\end{tabular}
\end{table}

\subsection{Structural descriptors at \texorpdfstring{\tauhalf}{tau=0.5}}

We extract 13 hand-crafted grid descriptors at trajectory fraction $\tau=0.5$ of the best trajectory,
which is the success path, or on failures the highest-scoring dead end. In other words, we read the
features from the first $\lceil n\times 0.5\rceil$ steps of an $n$-step trajectory. Object detection uses
4-connectivity, and the background color is the most frequent color in the input grid. Each base quantity
appears as a \emph{final} value (at the last step of the prefix), and where applicable as \emph{drift}
(final minus initial) and \emph{std} (standard deviation across prefix steps). The features are: object
count (\texttt{final}, \texttt{drift}, \texttt{std}); color entropy in bits (\texttt{final},
\texttt{drift}); number of distinct colors (\texttt{final}, \texttt{drift}); number of 4-connected
components (\texttt{final}, \texttt{drift}); horizontal and vertical symmetry (\texttt{final}); and
fragmentation, defined as $n_{\text{components}}/(H\times W)$ (\texttt{final}, \texttt{drift}).

The choice of $\tau=0.5$ was pre-registered before any analysis: it is the midpoint of the trajectory,
the point at which an early-stopping decision is most useful, and reading at a fixed fraction rather than
at the end avoids the score-saturation confound discussed in Section~\ref{sec:narrative}. A natural
question is how sensitive the signal is to it, which we answer by re-extracting the descriptors at
$\tau\in\{0.25, 0.5, 0.75\}$ from the stored step sequences (no solver re-runs) and repeating the identical
within-task analysis. Discriminability rises monotonically with $\tau$ (Table~\ref{tab:tau}): for Solver A
the within-task best-feature AUC goes from $0.841$ at $\tau=0.25$ to $0.885$ at $0.5$ to $0.945$ at $0.75$,
and for Solver B from $0.846$ to $0.884$ to $0.896$, all significant at $p<0.001$. Successful and failed
runs become easier to separate later in the trajectory, as expected. Reading at $\tau=0.5$ is therefore a
conservative choice that leaves roughly six points of AUC on the table for Solver A in exchange for an
earlier, more actionable readout; it is not tuned for maximum discriminability.

\begin{table}[H]
\centering
\caption{Temporal ablation: within-task best-feature AUC as a function of the readout fraction $\tau$,
pooled per solver. Confidence intervals are bootstrap; all cells significant at $p<0.001$ under the
within-task permutation null.}
\label{tab:tau}
\small
\begin{tabular}{lcc}
\toprule
$\tau$ & Solver A ($n=35$) & Solver B ($n=31$) \\
\midrule
0.25 & $0.841$ $[0.806, 0.877]$ & $0.846$ $[0.809, 0.882]$ \\
0.50 & $0.885$ $[0.849, 0.918]$ & $0.884$ $[0.849, 0.915]$ \\
0.75 & $0.945$ $[0.922, 0.967]$ & $0.896$ $[0.860, 0.928]$ \\
\bottomrule
\end{tabular}
\end{table}

\subsection{Analyses and metrics}

\paragraph{Within-task AUC.}
For each reliable diverse task, and for each feature, we compute the rank-based AUC separating successful runs
from failed ones. AUC is invariant to task-level affine shifts of the feature, so it measures purely
within-task discriminability. We report the per-task best-feature AUC averaged over tasks, with bootstrap
95\% confidence intervals ($n=2000$ task-level resamples) and a within-task label-permutation null (labels
shuffled within each task, $n=2000$).

\paragraph{Cross-solver transfer.}
For each ordered pair of (source, target) solver populations, we select the single best feature and its
predictive direction using only the source solver's diverse tasks. We then evaluate that frozen feature
blind on the target solver's held-out tasks. Because the feature and its direction are fixed on the
source, the target AUC is leakage-controlled.

\paragraph{Capacity de-confounding.}
We test the alternative that the features merely proxy solver budget in three ways: concordance on
config-contradictory pairs (pairs in which the succeeding run has strictly lower capacity); within-capacity-tier
AUC (within groups of equal-capacity configurations); and config-residualized AUC (the AUC of feature
residuals after regressing out the linear configuration-capacity effect via OLS).

\paragraph{Score and descriptor weak coupling.}
We test whether the descriptor signal merely restates the score trajectory by computing within-task
bidirectional $R^2$ between the score sequence $S$ and the descriptor vector $X$, together with a canonical
correlation analysis (CCA) between them.

\paragraph{Feature-space geometry.}
We summarize the effective dimensionality of the 13-D descriptor space by
$d_{\text{eff}} = \mathrm{tr}(C)^2/\mathrm{tr}(C^2)$ for covariance $C$, report the variance explained by
the first principal component, and re-run the within-task analysis after dropping the most collinear
feature pair as a robustness check.

\paragraph{Held-out endpoint.}
The pre-registered held-out test freezes the feature \texttt{n\_components\_final}, which was selected on
Stage 1 Solver A development tasks, and evaluates it on an assembled held-out pool of reliable diverse
tasks (40\% held out, stratified by $n_{\text{pos}}$, seed 1234). Significance uses the same within-task
label-permutation null, and we report a bootstrap 95\% CI.

Throughout, multiple-comparison correction uses Benjamini--Hochberg (BH).

\subsection{Pre-registered guardrails}

The following choices were fixed in advance and held identical across all stages and both solvers: the
score function $S(g,t)=\mathrm{matches}/(H\times W)$ with success at $S=1.0$; the readout fraction
$\tau=0.5$; all 13 feature definitions; the diversity criterion ($n_{\text{success}}\geq 1$ and
$n_{\text{failure}}\geq 1$); the reliability floor ($n_{\text{pos}}\geq 3$ and $n_{\text{neg}}\geq 3$);
the capacity formula ($\log_2$ term plus primitive penalty); and the seed-derivation scheme,
$\mathrm{MD5}(\texttt{f"\{42\}|\{task\_id\}|\{config\_id\}|\{seed\_idx\}"})[:8]$. No new features were
introduced after pre-registration, and the reliability floor was never relaxed.

\section{Results}

\subsection{How we arrived at the within-task signal}
\label{sec:narrative}

Before the individual results, we lay out the order in which we tested them, because that order is itself
part of the evidence. The convergent controls in this section were not assembled to support a conclusion
we had already reached. Each one was an attempt to explain the signal away, and we report them as the
explanations that did not hold.

At first, the natural worry was score saturation. The score trajectory flattens as a run approaches the
target, so any feature measured late in the trajectory might simply be reading off a near-solved grid. We
addressed this by reading features at the fixed fraction \tauhalf{} rather than at the end, and later by
checking the within-task relationship between the score sequence and the structural descriptors directly
(Section~\ref{sec:orthogonality}). The within-task coupling turned out to be negligible, so saturation did
not account for the signal.

The next worry was task identity. A feature that separates successes from failures across the whole
dataset may only be tracking which tasks are hard. We moved the entire analysis inside tasks: every AUC is
computed among runs on a single task, and significance is assessed by permuting outcome labels within each
task (Section~\ref{sec:within}). The signal survived this, which is the difference between a within-task
and a cross-task effect.

The third worry was solver capacity. More powerful configurations succeed more often and may also produce
distinctive grids, so the features could be a proxy for budget. We tested this three ways, including
residualizing out the configuration-capacity effect and measuring concordance on pairs where the
\emph{less} powerful configuration was the one that succeeded (Section~\ref{sec:confound}). The signal
survived all three.

The final and hardest worry was that the effect might be specific to one search algorithm. A within-solver
correlation, however well controlled, could still be a quirk of beam search. We answered this by freezing a
feature on one solver and evaluating it blind on a second, architecturally distinct solver
(Section~\ref{sec:cross}). It transferred in all four directions. That was the test we expected the signal
to fail, and it did not.

We present the rest of this section in roughly that order, from the base effect to the controls to the
cross-solver transfer.

\subsection{Within-task signal}
\label{sec:within}

\begin{figure}[H]
\centering
\includegraphics[width=\linewidth]{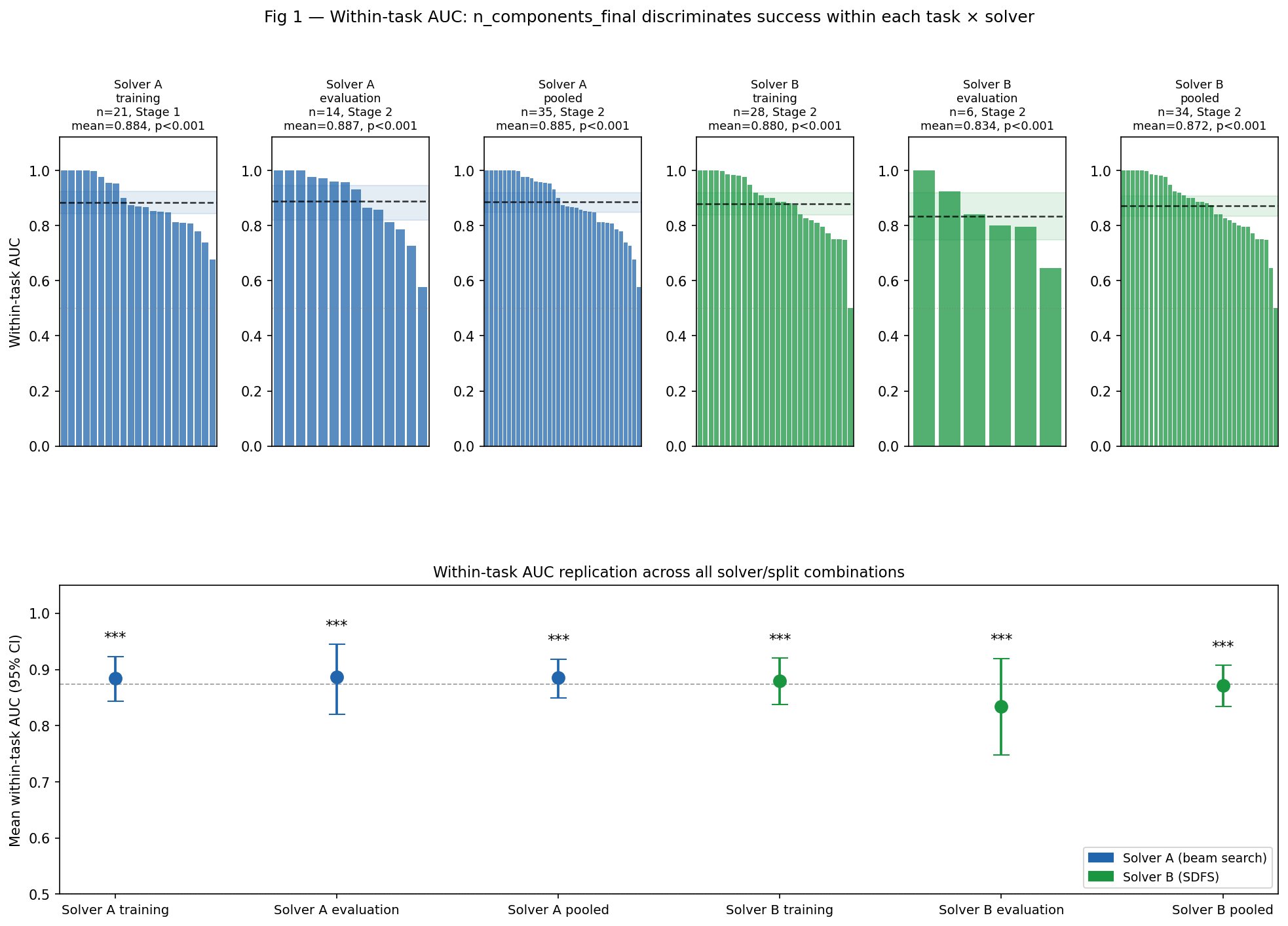}
\caption{Within-task best-feature AUC across solvers and splits. Top: per-task AUC for each of the six
solver/split populations, sorted within each panel, with the panel mean (dashed) and its bootstrap 95\%
band (shaded). Bottom: the same six means with 95\% confidence intervals on a common axis; all are
significant at $p<0.001$ (***). Solver B (SDFS, green) reproduces Solver A's within-task signal (beam
search, blue).}
\label{fig:within}
\end{figure}

Structural descriptors discriminate success from failure within the same task, and the effect replicates
across both solver architectures and both ARC splits (Table~\ref{tab:within}, Figure~\ref{fig:within}).
The pooled best-feature AUC is $0.885$ for Solver A and $0.872$ for Solver B, with $p<0.001$ in every cell
under the within-task label-permutation null. On Solver B pooled, 27 of 34 task-level tests survive BH
correction at $q<0.05$. The two solvers land on essentially the same number despite their different search
architectures, and their confidence intervals overlap. These analyses run over the reliable diverse tasks
only, which number 35 for Solver A and 34 for Solver B (Table~\ref{tab:data}); the 44{,}800-run total
provides coverage, but the effective sample for the within-task tests is on the order of 30 to 40 tasks per
solver, and we state this plainly so the figure is not mistaken for the analysis $n$.

The $0.885$ is a \emph{best-feature} statistic: for each task we take the most discriminative of the 13
descriptors and average that across tasks. Reporting the best of several features inflates a raw average,
which is exactly why significance is assessed against the within-task label-permutation null rather than
against $0.5$: the null is computed with the same best-of-13 selection on permuted labels, so the
selection advantage is present in the null too and the reported $p$ already accounts for it. The honest
unselected counterpart is the held-out AUC of $0.765$ in Section~\ref{sec:heldout}, where the feature is
frozen in advance with no per-task selection, and the gap between $0.885$ and $0.765$ is the cost of that
selection. For a descriptive sense of the typical feature rather than the best one, the mean within-task
AUC averaged across all 13 descriptors is $0.689$ for Solver A and $0.723$ for Solver B; the median-feature
figures are comparable. The average descriptor thus discriminates well above chance, while the best-feature
statistic carries a selection premium of $0.15$ to $0.20$ that the permutation null absorbs.

We also checked for fragile perfect-separation tasks, the kind where a within-task AUC of $1.0$ rests on
one or two points. Among the reliable diverse tasks of either solver, the number with AUC $=1.0$ on twelve
or fewer runs and only a single positive or negative is zero. The pre-registered reliability floor
($n_{\text{pos}}\geq 3$, $n_{\text{neg}}\geq 3$) excludes these cases by construction, so no task-level
perfect score in our results is driven by a single separating point.

A caution belongs here rather than only in the limitations. Some individual tasks reach a within-task AUC
of $1.0$ on as few as a dozen runs, sometimes with only a single success against the failures. A perfect
separation on that few points can occur by chance, so we do not read any individual task-level AUC as
evidence on its own. The claim rests on the distribution across tasks and on the within-task permutation
null, which is evaluated at the task level, not on the perfect-separation tasks taken individually. The
means we report are not inflated by these cases in any way that the permutation null does not already
account for.

\begin{table}[H]
\centering
\caption{Within-task best-feature AUC by solver and split. Confidence intervals are bootstrap
($n=2000$), and $p$-values are within-task label-permutation.}
\label{tab:within}
\small
\begin{tabular}{lrccc}
\toprule
Solver / Split & $N$ reliable tasks & Mean AUC & 95\% CI & Perm $p$ \\
\midrule
Solver A training   & 21 & 0.884 & $[0.843, 0.923]$ & $<0.001$ \\
Solver A evaluation & 14 & 0.887 & $[0.820, 0.945]$ & $<0.001$ \\
Solver A pooled     & 35 & \textbf{0.885} & $[0.849, 0.918]$ & $<0.001$ \\
Solver B training   & 28 & 0.880 & $[0.838, 0.920]$ & $<0.001$ \\
Solver B evaluation &  6 & 0.834 & $[0.747, 0.920]$ & $<0.001$ \\
Solver B pooled     & 34 & 0.872 & $[0.834, 0.907]$ & $<0.001$ \\
\bottomrule
\end{tabular}
\end{table}

\subsection{Cross-solver generalization}
\label{sec:cross}

\begin{figure}[H]
\centering
\includegraphics[width=\linewidth]{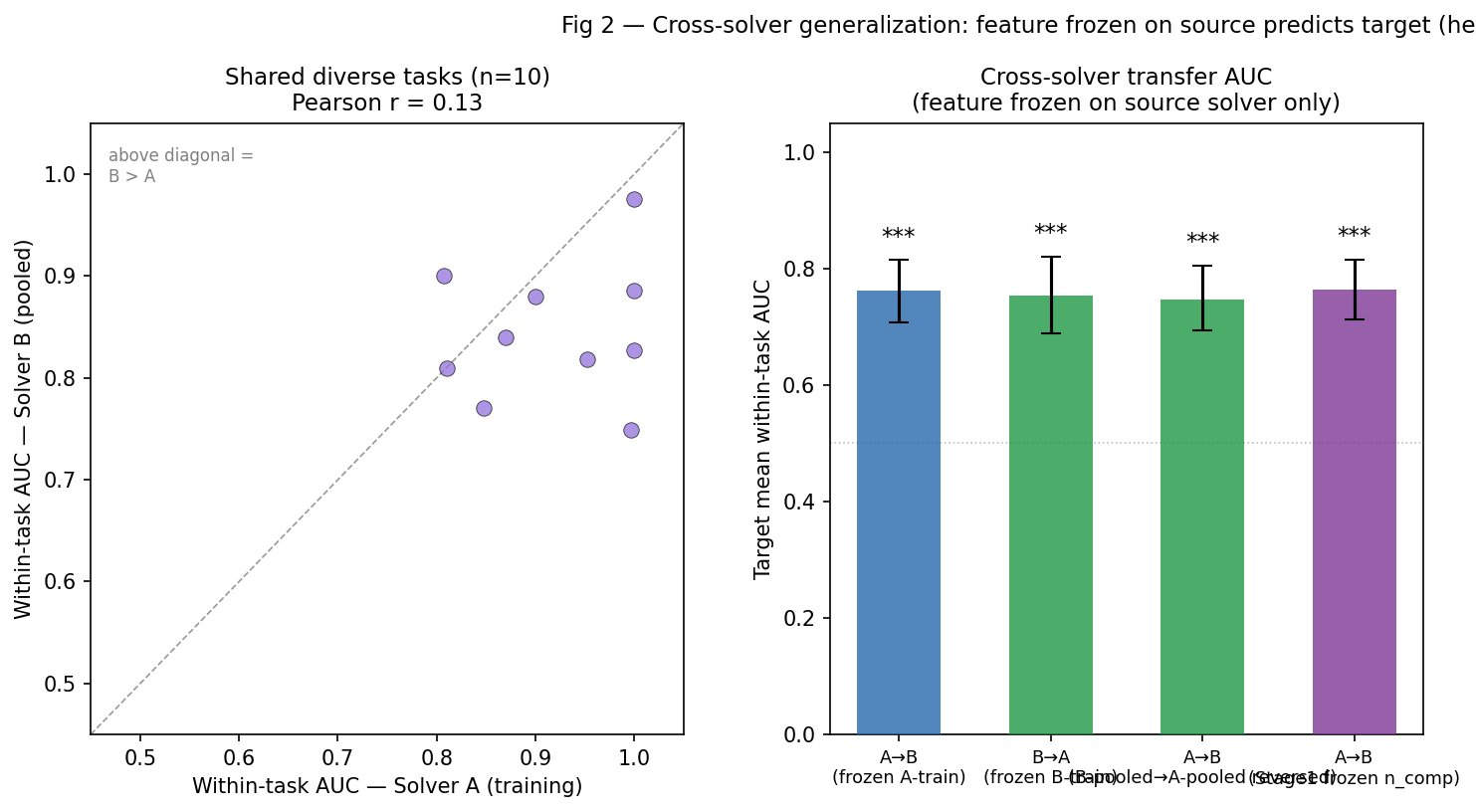}
\caption{Cross-solver generalization. Left: for tasks that are diverse under both solvers ($n=10$
shared), within-task AUC on Solver A versus Solver B. The low per-task correlation ($r=0.13$) shows that
transfer is not driven by a few shared easy tasks. Right: target mean within-task AUC for each transfer
direction, with the feature frozen on the source solver only; all four directions are significant at
$p<0.001$ (***), and the dotted line marks chance.}
\label{fig:cross}
\end{figure}

This pre-registered endpoint asks whether the signal is solver-specific, and the answer is that it is not.
Freezing the best feature and its direction on a source solver and evaluating blind on a target solver
yields significant transfer in all four directions (Table~\ref{tab:cross}, Figure~\ref{fig:cross}). Target
AUCs fall between $0.747$ and $0.762$, and every direction reaches $p<0.001$. Because the feature is
selected using only source-solver diverse tasks, this is leakage-controlled generalization across
architecturally distinct search strategies. The features that recur as frozen predictors are
\texttt{obj\_count\_final} (in the beam-to-SDFS directions) and \texttt{color\_entropy\_final} (in the
SDFS-to-beam directions).

\begin{table}[H]
\centering
\caption{Cross-solver transfer. The frozen feature is selected on the source solver
only, and the target AUC is evaluated blind. All four directions reach $p<0.001$.}
\label{tab:cross}
\small
\begin{tabular}{llccc}
\toprule
Direction & Frozen feature & Source AUC & Target AUC & 95\% CI \\
\midrule
A-training $\to$ B-pooled  & \texttt{obj\_count\_final}     & 0.786 & \textbf{0.762} & $[0.707, 0.815]$ \\
B-training $\to$ A-training & \texttt{color\_entropy\_final} & 0.783 & \textbf{0.753} & $[0.689, 0.820]$ \\
B-pooled $\to$ A-pooled    & \texttt{color\_entropy\_final} & 0.780 & \textbf{0.747} & $[0.694, 0.805]$ \\
A-pooled $\to$ B-pooled    & \texttt{obj\_count\_final}     & 0.774 & \textbf{0.762} & $[0.707, 0.815]$ \\
\bottomrule
\end{tabular}
\end{table}

As an additional pre-registered check, the Stage 1 frozen feature \texttt{n\_components\_final} (selected
on Solver A development tasks) evaluated blind on Solver B pooled gives AUC $=0.764$, 95\% CI
$[0.713, 0.815]$, $p<0.001$ ($n=34$ tasks).

\subsection{Held-out endpoint and its stability under clustering}
\label{sec:heldout}

\begin{figure}[H]
\centering
\includegraphics[width=\linewidth]{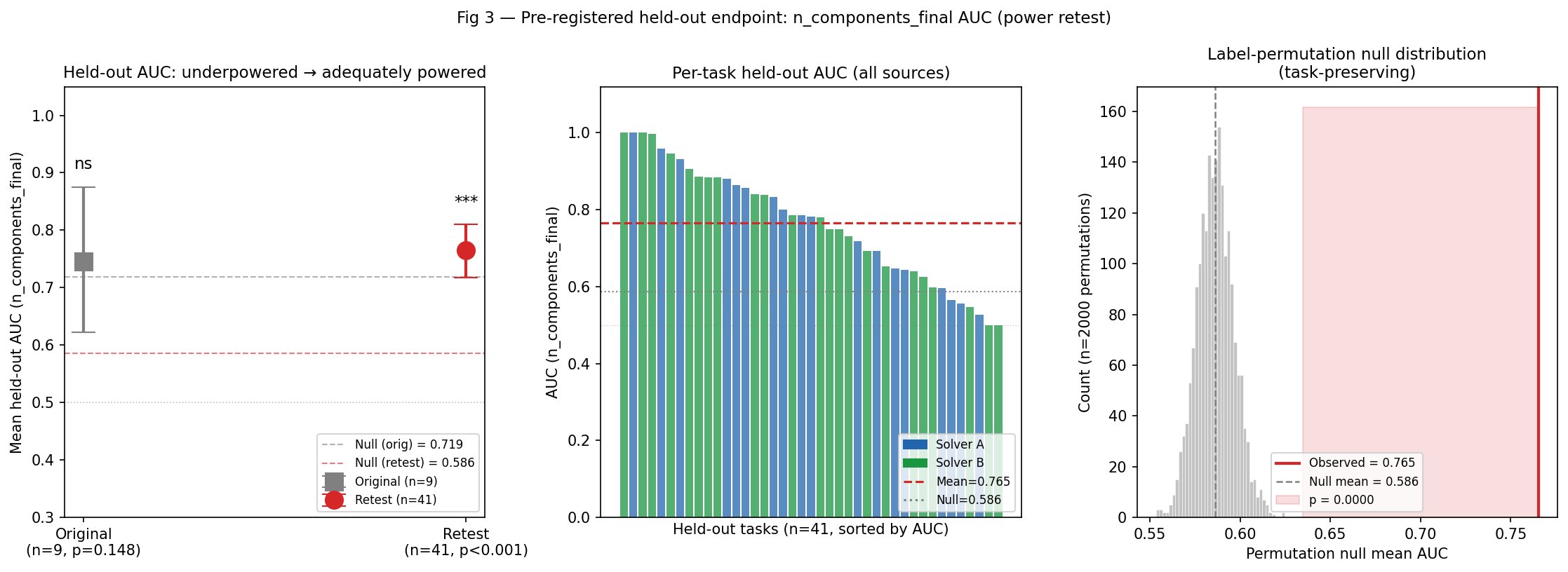}
\caption{Held-out endpoint for \texttt{n\_components\_final} on the 41 reliable diverse held-out entries.}
Left: per-task held-out AUC sorted by value, colored by source solver, with the pooled mean (dashed,
$0.765$) and the permutation-null mean (dotted, $0.586$). Right: the task-preserving label-permutation null
distribution ($n=2000$); the observed mean (solid red, $0.765$) lies far in the upper tail. 
\label{fig:heldout}
\end{figure}

We evaluate the pre-registered frozen feature \texttt{n\_components\_final}, with no re-selection, on the
assembled held-out pool of reliable diverse tasks. The pool draws from four sources outside the Stage 1
development split (Table~\ref{tab:heldout}, top), which yields 41 evaluable task-solver entries. The mean
held-out AUC is $0.765$ (95\% CI $[0.717, 0.810]$) against a label-permutation null mean of $0.586$, with
$p<0.001$ ($n=2000$ permutations). Figure~\ref{fig:heldout} shows the per-task spread and the permutation
null.

The 41 entries are not 41 independent draws, and we treat them accordingly. They come from 38 unique task
IDs, with 6 tasks appearing under both solvers, and within a task the runs share configuration structure.
To check that the result does not depend on treating these as independent, we recompute the endpoint under
resampling that respects the task structure. A cluster bootstrap that resamples whole tasks (keeping both
solver entries of a dual task together) gives a 95\% CI of $[0.723, 0.809]$, essentially unchanged from the
run-level interval, with no resample falling below chance. Collapsing the 6 dual-solver tasks to one entry
each leaves AUC $=0.775$ ($n=35$, $p<0.001$), and the more conservative step of dropping them entirely
leaves AUC $=0.788$ ($n=29$, $p<0.001$). (These counts fall below 38 and 35 because 3 held-out tasks have
an all-NaN \texttt{n\_components\_final} and are excluded.) The mean AUC rises slightly as the dependent
entries are removed, which is the opposite of what a sample-size artifact would produce. The held-out
result is therefore a property of the tasks, not an artifact of counting the same task twice. We report
what we have, 41 entries from 38 tasks, and we show it holds however the shared tasks are handled.

\begin{table}[H]
\centering
\caption{Held-out endpoint for the frozen feature \texttt{n\_components\_final}.}
\label{tab:heldout}
\small
\begin{tabular}{lr}
\toprule
Metric & Value \\
\midrule
Held-out task-solver entries evaluated & 41 \\
Mean held-out AUC & \textbf{0.765} \\
Bootstrap 95\% CI ($n=2000$) & $[0.717, 0.810]$ \\
Label-permutation null mean & 0.586 \\
Permutation $p$ ($n=2000$) & $<0.001$ \\
\midrule
\multicolumn{2}{l}{\emph{Per-source mean AUC}} \\
\quad Solver A evaluation ($n=14$)         & 0.777 \\
\quad Solver B training ($n=17$)           & 0.802 \\
\quad Solver A training held-out ($n=4$)   & 0.692 \\
\quad Solver B evaluation ($n=6$)          & 0.685 \\
\bottomrule
\end{tabular}
\end{table}

As noted above, the pool spans 38 unique task IDs, with 6 appearing under both solvers. We do not treat
those 6 as independent measurements. Section~\ref{sec:heldout} reports the endpoint with them collapsed to
one entry each and with them dropped, and the result is significant in both cases, so the way they are
counted does not carry the conclusion.

\subsection{The signal is not a capacity proxy}
\label{sec:confound}

\begin{figure}[H]
\centering
\includegraphics[width=0.92\linewidth]{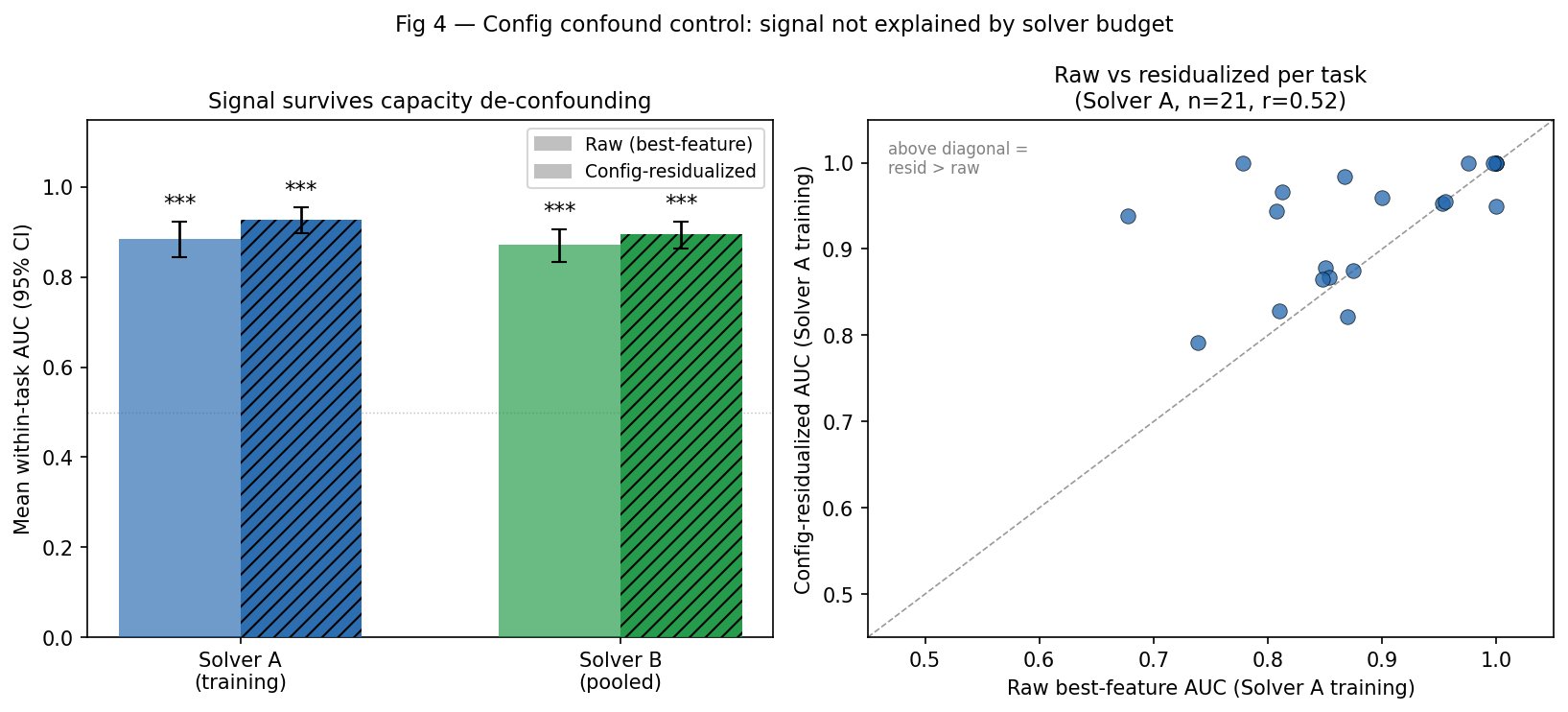}
\caption{Configuration confound control. Left: raw best-feature AUC versus configuration-residualized AUC
for both solvers, with 95\% confidence intervals; both stay well above chance after the linear capacity
effect is removed ($p<0.001$, ***). Right: per-task raw versus residualized AUC on Solver A ($n=21$,
$r=0.52$); most tasks sit on or above the diagonal, so residualization does not erode the signal.}
\label{fig:confound}
\end{figure}

A natural alternative is that high-capacity configurations produce both distinctive feature values and
higher success rates, which would make the features a proxy for solver budget. Three analyses rule this
out (Figure~\ref{fig:confound}).

At first, consider the config-contradictory pairs: 545 pairs in which the succeeding run has strictly
\emph{lower} capacity. Across these pairs, 10 of 13 features achieve concordance above $0.65$, correctly
ranking the successful run even though it had less budget, with \texttt{n\_colors\_drift} reaching $0.97$
and \texttt{n\_colors\_final} $0.88$. Next, the mean within-capacity-tier AUC is $0.948$ (over 59 valid
task-tier groups), so the features discriminate within groups of equally powerful configurations. Finally,
after regressing out the linear configuration-capacity effect, the configuration-residualized AUC remains
high and significant: $0.927$ on Solver A training and $0.896$ on Solver B pooled, both $p<0.001$, with a
34-task mean of $0.923$ across the pooled diverse set. The signal is not explained by solver capacity.

\subsection{Score and descriptor weak coupling}
\label{sec:orthogonality}

The predictive signal does not appear to be a restatement of the score trajectory. Within tasks, the score
sequence $S$ and the descriptor vector $X$ show no detectable linear relationship: bidirectional within-task
$R^2$ is effectively zero ($R^2(S\to X)\approx -0.02$, $R^2(X\to S)\approx -0.03$, where the small negative
values reflect cross-validated $R^2$ on a relationship indistinguishable from none). We describe this as
weak coupling rather than independence, since these statistics rule out a linear within-task relationship
but not every form of dependence. The two share only a cross-task difficulty axis, captured by a first
canonical correlation of $r=0.585$ that vanishes within tasks. To the resolution these tests provide, the
within-task descriptor signal appears to carry information the score does not, though we cannot rule out
nonlinear dependence.

\subsection{Feature-space geometry}
\label{sec:geometry}

The 13 descriptors do not contribute 13 independent signals, and it is important to say so plainly rather
than let the count imply otherwise. The descriptor space is near rank-1: $d_{\text{eff}}=1.069$, and the
first principal component explains $96.7\%$ of the variance. Two features are nearly identical
($\mathrm{corr}(\texttt{obj\_count\_final}, \texttt{n\_components\_final})=0.996$), and most of the
predictive content lives along a single axis that is best read as overall grid complexity, roughly the
number of objects or connected components present. The honest characterization of the result is therefore
not ``thirteen descriptors predict success'' but ``a one-dimensional complexity descriptor predicts
success, and twelve of our hand-crafted features are largely restatements of it.''

\begin{figure}[H]
\centering
\includegraphics[width=0.82\linewidth]{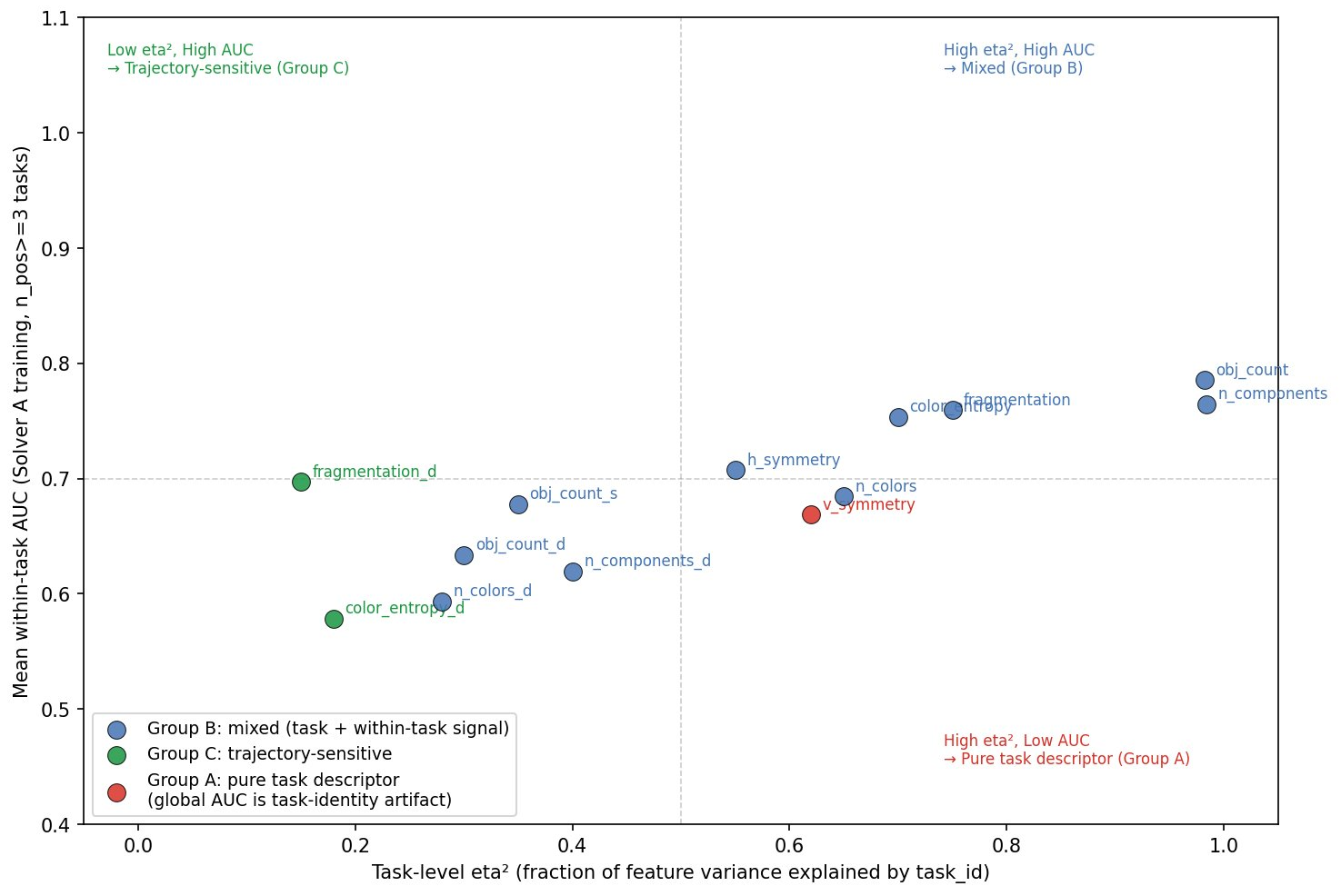}
\caption{Feature taxonomy: task-level $\eta^2$ (fraction of feature variance explained by task identity)
against mean within-task AUC. Group A (red, \texttt{v\_symmetry\_final}) has high $\eta^2$ but
near-chance within-task AUC, so its apparent global predictiveness is a task-identity artifact. Group B
(blue) carries both task-level and within-task signal, and includes the strongest predictors
(\texttt{obj\_count\_final}, \texttt{n\_components\_final}, color entropy). Group C (green) is
trajectory-sensitive, with low $\eta^2$ but useful within-task AUC.}
\label{fig:taxonomy}
\end{figure}

We think this is a clarification of the finding rather than a retreat from it. A single interpretable scalar
that predicts within-task success, transfers across two solver architectures, and survives the controls in
this paper is a cleaner and more falsifiable claim than an entangled thirteen-feature classifier would be.
It also matches the cross-solver result, where the transferable predictors are exactly the complexity
counts (\texttt{obj\_count\_final}, \texttt{n\_components\_final}) rather than the more exotic descriptors.
The signal is robust to the redundancy: dropping the two most collinear features leaves a within-task AUC
of $0.865$ ($p<0.001$), so the axis is not an artifact of double-counting one quantity.
Figure~\ref{fig:taxonomy} organizes the 13 features by how much of their variance is explained by task
identity ($\eta^2$) against their within-task AUC, which separates the features that are essentially task
labels from those that track the trajectory.

We confirmed this directly with a multivariate model. A logistic regression on all 13 descriptors reaches
an AUC of $0.756$, slightly \emph{below} the $0.765$ of the single frozen feature, so combining the
descriptors adds nothing over the one complexity axis and in fact loses a little to the extra parameters.
A leave-one-feature-out sweep is consistent with this: \texttt{fragmentation\_final} is the most load-bearing
single descriptor (AUC drops $0.040$ when it is removed), followed by \texttt{n\_colors\_final} and
\texttt{obj\_count\_final} (each about $0.024$), while dropping a redundant feature such as
\texttt{color\_entropy\_final} or \texttt{obj\_count\_std} slightly \emph{improves} the model, the standard
signature of multicollinearity. Nothing meaningful appears to live off the dominant axis.

\section{Application: Early-Stopping}
\label{sec:earlystop}

\begin{figure}[H]
\centering
\includegraphics[width=\linewidth]{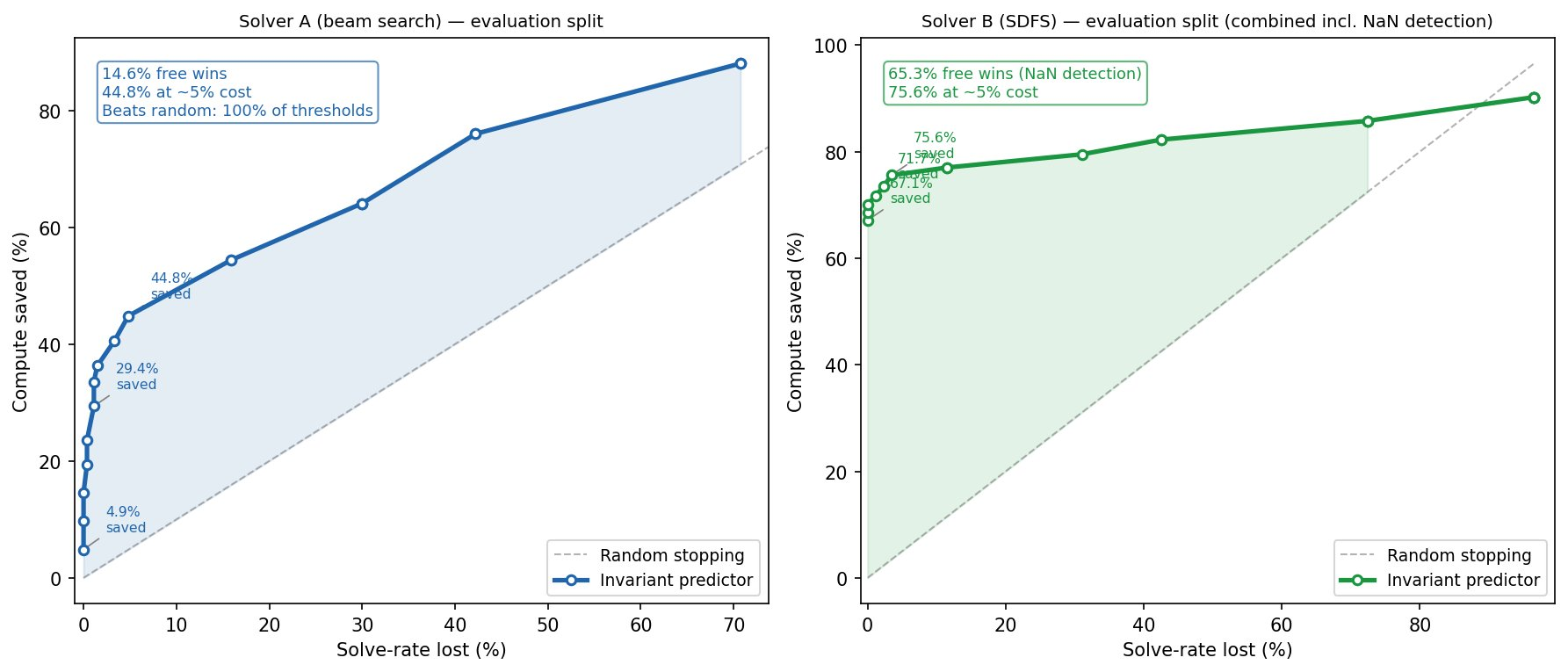}
\caption{Compute and solve-rate Pareto frontiers for early-stopping at \tauhalf{}. The dashed diagonal is
the random-stopping baseline, where compute saved equals solve-rate lost. Left: Solver A (beam search),
where the descriptor-based predictor lies well above the diagonal at every threshold (14.6\% free wins, 44.8\%
saved at $\sim$5\% cost, beats random at 100\% of thresholds). Right: Solver B (SDFS), combined curve
including NaN-detection, where the savings are dominated by the zero-cost 65.3\% NaN free wins.}
\label{fig:pareto}
\end{figure}

If structural descriptors at \tauhalf{} predict failure, then they can terminate doomed runs early. We
simulate this post-hoc on the held-out ARC evaluation split, using the cross-solver frozen feature
\texttt{obj\_count\_final}, where a higher value means a predicted failure. The feature was frozen on
Solver A development tasks, so no Solver B data informs the predictor. As a compute proxy we use
$\text{beam width}\times\text{max depth}$ for Solver A and \texttt{max\_nodes} for Solver B. Absolute
percentages carry proxy uncertainty because raw per-step counts were not stored, but the relative Pareto
comparison against a random-stopping baseline is valid. The two solvers behave differently, so we report
them separately (Figure~\ref{fig:pareto}).

\paragraph{Solver A, beam search.}
The predictor yields a strong compute and solve-rate Pareto frontier (Table~\ref{tab:earlyA}). Stopping
runs whose \texttt{obj\_count\_final} exceeds a free-win threshold saves $14.6\%$ of compute at zero
solve-rate cost. Accepting a $\sim$1\% solve-rate cost saves $33.6\%$, and accepting $\sim$5\% saves
$44.8\%$. The predictor beats random stopping at $100\%$ of threshold points (median gap $33.8$ percentage
points, peak $40.0$). At the median threshold, 5{,}057 of 11{,}200 runs are stopped while only 13 of 270
successes are lost, against roughly 122 expected under random stopping of the same fraction, which is
about a tenfold improvement.

\begin{table}[H]
\centering
\caption{Solver A (beam search) early-stopping Pareto points on the held-out ARC evaluation split.}
\label{tab:earlyA}
\small
\begin{tabular}{lcc}
\toprule
Operating point & Compute saved & Solve-rate retained \\
\midrule
Free wins (0\% cost)      & \textbf{14.6\%} & 100\% \\
$\sim$1\% solve-rate cost & \textbf{33.6\%} & 98.9\% \\
$\sim$5\% solve-rate cost & \textbf{44.8\%} & 95.2\% \\
\bottomrule
\end{tabular}
\end{table}

\paragraph{Solver B, SDFS.}
On SDFS the dominant savings come from a different source. $65.7\%$ of Solver B evaluation runs (7{,}357 of
11{,}200) have no valid features at \tauhalf{}, and all of them are failures arising from the 229
degenerate tasks characterized in Section~\ref{sec:mechanism}. Detecting these NaN-feature runs and
stopping them saves $65.3\%$ of SDFS compute at zero solve-rate cost. The \texttt{obj\_count\_final}
predictor applied to the remaining non-NaN runs adds about $10.2\%$ further savings at a $\sim$3.4\%
solve-rate cost, but it beats random at only $42\%$ of thresholds (peak gap $6.8$ pp), because the non-NaN
survivors are harder borderline cases. Combining NaN-detection with the predictor saves $75.6\%$ of SDFS
compute at a $\sim$5\% solve-rate cost. We stress that the large SDFS figure is driven by NaN-detection, a
structural property of the solver and DSL pairing, not by predictor discriminability, which is why we
present the two solvers separately.

\section{Mechanism: DSL Coverage Gap}
\label{sec:mechanism}

\begin{figure}[H]
\centering
\includegraphics[width=\linewidth]{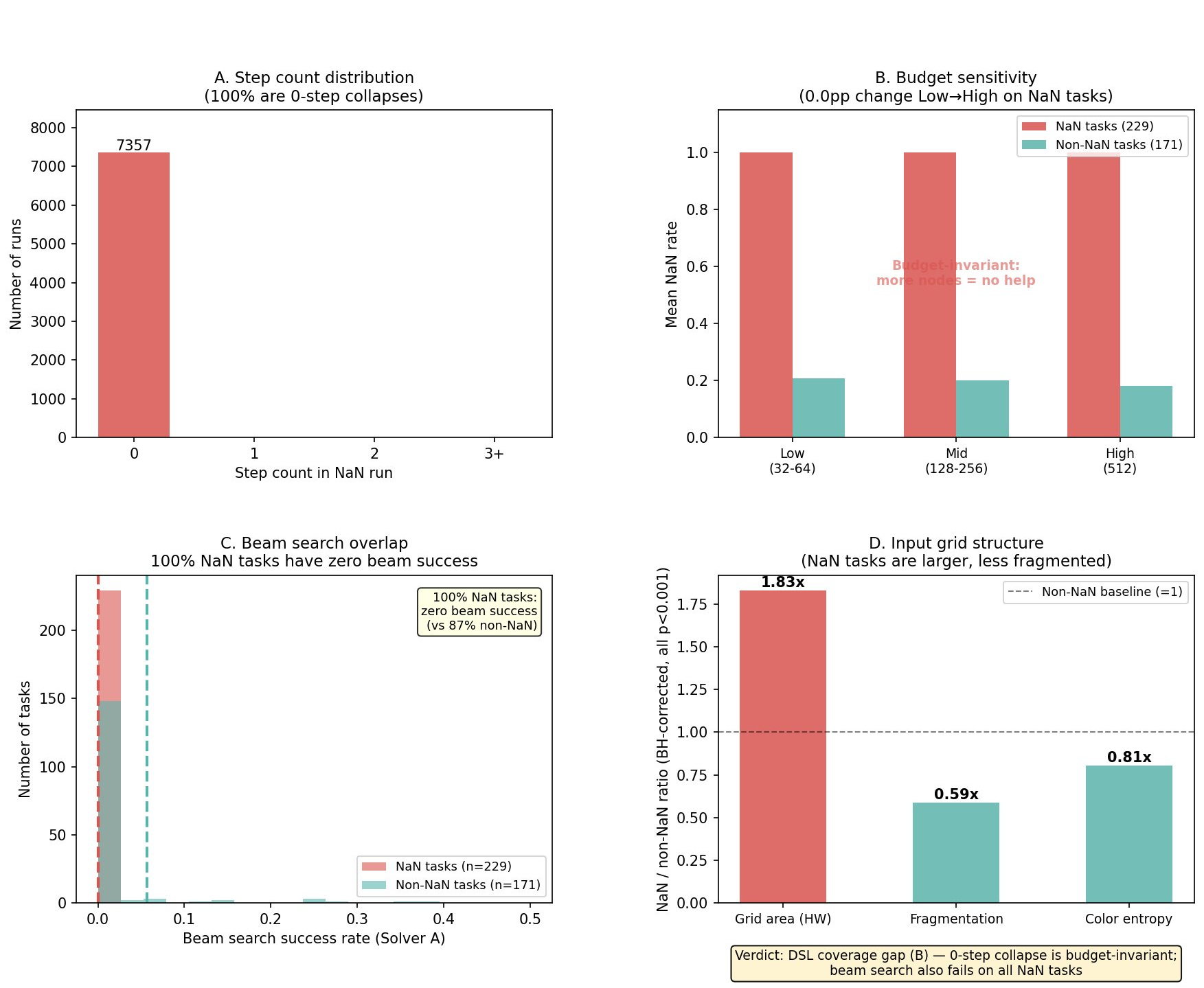}
\caption{Mechanism of the degenerate SDFS trajectories on 229 of 400 evaluation tasks. (A) All 7{,}357 NaN
runs have exactly 0 steps. (B) The NaN rate is flat across node-budget tiers (32 to 512), so more budget
does not help. (C) Every NaN task has zero beam-search success, versus 87\% zero-success for non-NaN
tasks, so the difficulty is task-level rather than SDFS-specific. (D) NaN tasks are structurally distinct,
with larger grids ($1.83\times$ area) and lower fragmentation ($0.59\times$) and color entropy
($0.81\times$). Together these are consistent with a DSL coverage gap.}
\label{fig:nan}
\end{figure}

The degenerate SDFS trajectories behind the NaN-detection savings have a consistent explanation. On 229 of
400 ARC evaluation tasks, SDFS produces 0-step trajectories, with no valid successor from the initial grid,
under all 28 configurations. Four converging analyses (Table~\ref{tab:nan}, Figure~\ref{fig:nan}) are
consistent with a DSL coverage gap, and weigh against budget exhaustion, a depth requirement, or an
SDFS-specific artifact.

\begin{table}[H]
\centering
\caption{Degenerate SDFS trajectories. Four analyses are jointly consistent with a DSL coverage gap.}
\label{tab:nan}
\small
\begin{tabular}{p{0.24\linewidth}p{0.42\linewidth}p{0.26\linewidth}}
\toprule
Analysis & Key result & Implication \\
\midrule
Step count & 100\% of the 7{,}357 NaN runs have exactly 0 steps
& No primitive produces a valid successor, a coverage failure \\
Budget sensitivity & NaN rate $=1.000$ at every tier (32--64, 128--256, 512 nodes), a $0.0$ pp change
& Rules out budget exhaustion and depth requirement \\
Beam-search overlap & 100\% of the 229 NaN tasks have zero beam-search success (vs.\ 86.5\% for non-NaN)
& Task-level, not solver-specific; beam search also fails \\
Input grid structure & NaN tasks have larger grids ($H\times W$: 296 vs.\ 162 cells, rank-biserial
$0.418$, $p<0.001$), lower fragmentation and color entropy; 8 of 11 properties differ after BH
& Grid size correlates with difficulty, not a direct mechanism \\
\bottomrule
\end{tabular}
\end{table}

The most telling evidence is the combination of 0-step collapse and budget invariance. If the constraint
were node budget, then some 512-node runs would take at least one step, and none do. The most consistent
reading is that the primitive library does not produce any valid transition from these input grids,
regardless of search strategy or compute, and the same tasks defeat beam search universally. We therefore treat NaN-feature detection as a
zero-cost early-stopping signal (Section~\ref{sec:earlystop}). One sub-question remains open and
non-blocking: whether the step-0 failure reflects no-op primitives, where the output equals the input, or
primitives that raise exceptions or return non-grid types at step 0. Distinguishing the two would require
verbose primitive-level logging on 10 to 15 NaN tasks, and the coverage-gap framing holds either way.

\section{Discussion}

These results support the claim that structural properties of intermediate ARC grids carry within-task
information about symbolic-solver success, and, the central point, that this information is not tied to one
search algorithm. A feature frozen on beam search predicts success on a depth-first solver, and the reverse
holds as well, at AUC near $0.75$ in every direction. The within-task framing, the leakage-controlled
cross-solver transfer, the capacity de-confounding, and the weak score coupling together argue against the
obvious confounds of task identity, solver budget, and the score trajectory. Taken together they point to
the signal being a property of how symbolic search over this DSL traverses grid space on ARC, rather than
an idiosyncrasy of one solver.

Two features recur as transferable predictors, \texttt{obj\_count\_final} and
\texttt{color\_entropy\_final}, and the near rank-1 geometry of the descriptor space ($d_{\text{eff}}=1.069$,
PC1 $=96.7\%$) shows that much of the predictive content sits in a small number of correlated
structural-complexity counts. This fits a simple reading: runs whose intermediate grids stay
over-fragmented or over-complex at the halfway point are less likely to converge to the exact target. The
early-stopping application makes that reading concrete and useful, saving a third of beam-search compute at
a $\sim$1\% solve-rate cost, an order of magnitude better than random. The DSL coverage analysis then turns
a confounding ``no-features'' failure mode into a principled, zero-cost stopping signal while pointing to a
likely limitation of the primitive library.

We are deliberate about what this work does and does not claim. It is an empirical characterization of
search dynamics, not a new ARC solver. The contribution is to show that intermediate grid structure carries
transferable information about symbolic-search success, and to quantify how much of that information
survives controls for task identity, solver capacity, and the score. We do not show an improvement in
ARC \emph{solve rate}, which is the metric the field cares about most. We tested the natural way one might
hope to get there. If the descriptor reliably flags doomed runs, then under a fixed compute budget one
could terminate those runs early and reallocate the freed budget toward additional restarts on unsolved
tasks, which would convert the compute saving into extra solves. We simulated exactly this on the held-out
evaluation split, using the frozen \texttt{obj\_count\_final} predictor to terminate runs at \tauhalf{} and
re-drawing already-computed runs to model the reallocation, with a random-reallocation control spending the
same freed budget. The result is a clear negative: descriptor-guided reallocation does not reliably raise
the solve count, averaging $+0.2$ solves across budget levels against $-1.0$ for random reallocation, with
the one positive point ($+2$ at 21 configurations per task) matched exactly by the random control. At the
largest budget the descriptor policy is slightly worse than baseline, because with no unsolved capacity
left to reallocate to, terminating a borderline run occasionally cuts off one that would have succeeded.

We report this negative result plainly because it bounds the practical claim. The early-stopping result is
a compute saving on symbolic baselines, and within our data that saving cannot be turned into a higher
solve rate. A signal that is informative about an individual run's outcome is not, on its own, enough to
make a solver solve more tasks: doing so would require either a budget regime with genuine unsolved
capacity to reallocate into, or a richer signal than a single complexity axis. What we establish is the
prerequisite and its current ceiling, namely that a cheap, interpretable descriptor measured halfway
through a run is informative about its outcome and transfers across solver architectures, but that this
informativeness does not by itself yield more solutions under a fixed budget.

\section{Limitations}
\label{sec:limitations}

\paragraph{Held-out pool and its structure.}
The held-out endpoint rests on 41 reliable diverse entries from 38 unique tasks, all drawn from sources
outside the development split on which the feature was frozen, and all satisfying the pre-registered
reliability floor ($n_{\text{pos}}\geq 3$, $n_{\text{neg}}\geq 3$), which we did not relax. Six tasks
contribute an entry under each solver. As reported in Section~\ref{sec:heldout}, the endpoint is
significant under a task-clustered bootstrap (95\% CI $[0.723, 0.809]$) and remains significant whether
those six are collapsed to one entry each (AUC $=0.775$, $n=35$) or dropped entirely (AUC $=0.788$,
$n=29$), so the conclusion does not rest on treating them as independent. The honest scope of the claim is
this set of tasks: we show the frozen feature predicts success on held-out tasks it was never tuned on,
across both solvers, and that the result is stable to how the shared tasks are counted.

\paragraph{One dominant axis, and simple features.}
As Section~\ref{sec:geometry} states, the predictive content is largely one-dimensional and the strongest
features (\texttt{n\_components\_final}, \texttt{obj\_count\_final}) are simple complexity counts, not deep
transformation-specific quantities. A reader who expected a rich multi-feature signal will not find one
here. We regard the single-axis nature as the actual result rather than a defect, but it does bound the
claim: what we show is that a grid-complexity descriptor predicts success, not that an elaborate family of
structural features does.

\paragraph{Compute proxy.}
The early-stopping savings use a compute proxy ($\text{beam width}\times\text{max depth}$ for Solver A and
\texttt{max\_nodes} for Solver B), because raw per-step counts were not stored. Relative comparisons
against the random baseline are valid, but absolute saving percentages carry proxy uncertainty. For
successful Solver A runs, the proxy is an upper bound, making the reported savings slightly
conservative.

\paragraph{Solver B early-stopping.}
The large SDFS compute savings are dominated by NaN-detection of degenerate runs, which reflects the DSL
coverage gap rather than predictor discriminability. On the non-degenerate SDFS runs the predictor beats
random at only $42\%$ of thresholds. The strong, general early-stopping claim is specific to Solver A.

\paragraph{Scope.}
All results concern symbolic search over a single \texttt{arc-dsl} primitive library, scored by exact
pixel match. Whether the same structural signal transfers to neural or hybrid ARC solvers, or to other
DSLs, is untested.

\paragraph{Impact on solving.}
The practical contribution is a compute saving on symbolic baselines, not an improvement in solve rate or
generalization, which are the metrics the ARC community weighs most heavily. We tested whether the saving
could be converted into solves by reallocating freed budget (Discussion), and found it could not within our
data: descriptor-guided reallocation was statistically indistinguishable from random reallocation. Readers
evaluating the work for its bearing on ARC reasoning should treat it as a characterization of search
dynamics, with a demonstrated compute application and a demonstrated negative on the solve-rate application,
rather than as a step toward higher scores.

\paragraph{No mechanistic theory.}
We report an empirical regularity without a model of why it holds. We can describe the pattern, that runs
whose grids stay over-complex at the halfway point are less likely to converge, but we do not have a theory
that predicts which descriptor should matter or by how much. The interpretation in terms of grid complexity
is a post-hoc reading of the data, not a derivation.

\section{Conclusion}

Measured at the halfway point of a search trajectory, a structural grid descriptor separates successful
from failed ARC solver runs within the same task. Most of the signal lies along a single grid-complexity
axis, it generalizes across two architecturally distinct solvers, and it withstands controls for task
identity, solver capacity, and the score trajectory. One use of it, early-stopping, saves about a third of
beam-search compute at near-zero solve-rate cost. Along the way, we characterize a failure mode consistent
with a DSL coverage gap, where 229 of 400 evaluation tasks resist both solvers and where the same signal
doubles as a zero-cost stopping rule. We present this as an empirical characterization of symbolic-search
dynamics on ARC: a cheap, interpretable signal in intermediate grid structure is informative about a run's
outcome, consistently enough to transfer across solvers. Whether that signal can be turned into higher
solve rates, rather than cheaper search, is tested directly: reallocating the freed budget did not beat
random reallocation, so, in our setting, the signal lowers the cost of search without raising how much it
solves. Closing that gap is what we think the result opens.

\end{document}